\documentclass[a4paper]{article}
\usepackage{INTERSPEECH2022}
\usepackage{booktabs}
\usepackage{balance}

\usepackage{amsmath,epsfig,amssymb,amsthm,url,amsfonts}
\usepackage{algorithm}
\usepackage{algpseudocode}
\usepackage{multirow}
\usepackage{mdframed}
\usepackage{url}
\usepackage{enumitem}
\usepackage[makeroom]{cancel}
\usepackage{dblfloatfix}
\usepackage{array}
\usepackage{comment}
\usepackage{epstopdf}
\usepackage{float}
\usepackage{subfig}
\usepackage{breqn}
\usepackage{cite}
\usepackage{xcolor}
\usepackage{cases}
\usepackage{tabularx}
\usepackage[printonlyused,withpage]{acronym}
\usepackage{balance}
\usepackage{graphbox} 

\setlist[itemize]{label=$\triangleright$}
\newtheoremstyle{break}
{}
{}
{\itshape}
{}
{\bfseries}
{.}
{\newline}
{}
\theoremstyle{break}

\theoremstyle{definition}

\newcommand{\vect}[1]{\mathbf{#1}}
\newcommand{\bs}[1]{\boldsymbol{#1}}
\newcommand{\E}{\mathbb{E}}
\usepackage{url}

\makeatletter
\def\thmhead@plain#1#2#3{%
	\thmname{#1}\thmnumber{\@ifnotempty{#1}{ }\@upn{#2}}%
	\thmnote{ {\the\thm@notefont#3}}}
\let\thmhead\thmhead@plain
\makeatother

\graphicspath{{./figs/}}

\newcommand{\argmax}{\operatornamewithlimits{argmax}}
\newcommand{\argmin}{\operatornamewithlimits{argmin}}

\newcommand{\Rbb}{{\mathbb{R}}}

\newcommand{\diag}{\mbox{{diag}}}

\newsavebox\mybox

\acrodef{SE}{speech enhancement}
\acrodef{STFT}{short-time Fourier transform}
\acrodef{STOI}{short-time objective intelligibility}
\acrodef{PSD}{power spectral density}
\acrodef{NMF}{nonnegative matrix factorization}
\acrodef{AV}{audio-visual}
\acrodef{DNN}{deep neural network}
\acrodefplural{DNNs}{deep neural networks}
\acrodef{VAE}{variational auto-encoder}
\acrodefplural{VAEs}{variational auto-encoders}
\acrodef{CVAE}{conditional variational auto-encoder}
\acrodefplural{CVAEs}{conditional variational auto-encoders}
\acrodef{A-VAE}{audio VAE}
\acrodef{V-VAE}{visual VAE}
\acrodef{AV-CVAE}{audio-visual CVAE}
\acrodef{ROI}{region of interest}
\acrodef{MCMC}{Markov Chain Monte Carlo}
\acrodef{EM}{expectation-maximization}
\acrodef{MCEM}{Monte Carlo expectation-maximization}
\acrodef{TF}{time frequency}
\acrodef{ELBO}{evidence lower bound}
\acrodef{ROI}{region of interest}
\acrodef{LR}{Living Room}
\acrodef{SDR}{signal-to-distortion ratio}
\acrodef{PESQ}{perceptual evaluation of speech quality}
\acrodef{ASE}{audio speech enhancement}
\acrodef{VSE}{visual speech enhancement}
\acrodef{AVSE}{audio-visual speech enhancement}
\acrodef{SNR}{signal-to-noise ratio}
\acrodefplural{SNRs}{signal-to-noise ratios}
\acrodef{LSTM}{long short-term memory}
\acrodef{DNNs}{deep neural networks}

\newcommand{\compresslist}{
	\setlength{\itemsep}{1pt}
	\setlength{\parskip}{0pt}
	\setlength{\parsep}{0pt}
}

\title{A Sparsity-promoting Dictionary Model for Variational Autoencoders}

\name{Mostafa Sadeghi, Paul Magron}
\address{Université de Lorraine, CNRS, Inria, LORIA, F-54000 Nancy, France}
\email{mostafa.sadeghi@inria.fr, paul.magron@inria.fr}

\begin{document}

\maketitle
\begin{abstract}
Structuring the latent space in probabilistic deep generative models, e.g., variational autoencoders (VAEs), is important to yield more expressive models and interpretable representations, and to avoid overfitting. One way to achieve this objective is to impose a sparsity constraint on the latent variables, e.g., via a Laplace prior. However, such approaches usually complicate the training phase, and they sacrifice the reconstruction quality to promote sparsity. In this paper, we propose a simple yet effective methodology to structure the latent space via a sparsity-promoting dictionary model, which assumes that each latent code can be written as a sparse linear combination of a dictionary's columns. In particular, we leverage a computationally efficient and tuning-free method, which relies on a zero-mean Gaussian latent prior with learnable variances.  We derive a variational inference scheme to train the model. Experiments on speech generative modeling demonstrate the advantage of the proposed approach over competing techniques, since it promotes sparsity while not deteriorating the output speech quality.
\end{abstract}


\section{Introduction}\label{sec:intro}

Unsupervised representation learning~\cite{Bengio2013replearning} is defined as the task of automatically extracting useful information from unlabeled data, in the form of a \emph{feature} or \emph{representation} vector, which can be subsequently used in downstream tasks. To that end, one successful approach, which has gained much attention in the past years, is based on variational autoencoders (VAEs)~\cite{KingW14}. These models explicitly consider a latent vector which encapsulates some information about the data. A VAE model consists of a stochastic \emph{encoder} (a recognition network), which transforms the input data into a latent space whose dimension is usually much lower than the original data space, and a stochastic \emph{decoder} (a generative network), which produces data back in the original space from the encoded latent representation. VAEs have been successfully exploited in a variety of speech processing related tasks, such as speech enhancement~\cite{leglaive2018variance}, source separation~\cite{nguyen2021deep}, speech recognition~\cite{Tan2016asr}, and speech generation and transformation~\cite{Hsu2017LearningLR}.

A common practice in a VAE framework consists in assuming a standard normal distribution as the prior distribution of the latent space. However, this is not effective, as it might not efficiently capture the underlying distribution of the data. Structuring the latent space by imposing some meaningful constraints and distributions is thus of paramount importance to obtain more expressive and interpretable representations. A promising approach is to favor parsimonious or \emph{sparse} latent representations, that is, feature vectors with mostly zeroes or small magnitude entries. Indeed, sparsity is known to promote a variety of benefits, such as increasing interpretability of the latent features and reducing the risk of overfitting~\cite{Asperti2019sparsevae}. It has proven very successful for feature learning in many fields such as computer vision~\cite{Gregor2015images} or natural language processing~\cite{Prokhorov2020nlp}.

In this regard, a sparse prior has been proposed in \cite{mathieu2019disentangling} in the form of a mixture of two Gaussian distributions with fixed variances, where one of them is very small in order to encourage sparsity. A mixing parameter balances the contribution of the two distributions, and thus the amount of sparsity. Tonolini \textit{et al.} \cite{tonolini2020variational} proposed a variational sparse coding (VSC) framework, where they consider a Spike-and-Slab distribution~\cite{Andersen2014spikeandslab} for the encoder, which is a mixture of a Gaussian distribution, as in a standard VAE, and a sparsity-promoting Delta function. The prior distribution has the same form as the encoder but instead of directly processing the original data as input, some (unknown) pseudo-input data are defined and learned jointly with the VAE parameters. The sparsity level is tuned via a user-defined parameter, which governs the prior distribution of the mixing variable. Following a different strategy,~\cite{miao2021incorporating} proposed a \textit{deterministic} dimension selector function, so that some dimensions of the latent vector are deactivated via a point-wise multiplication. The amount of sparsity is then monitored via an entropy-based regularization term. Similarly, in \cite{moran2021identifiable} a \textit{stochastic} per-feature masking variable is applied to the latent representation via point-wise multiplication to zero out some dimensions of the latent vector. A hierarchical Spike-and-Slab Lasso prior is assumed for the masking variable, consisting of two Laplace distributions with different parameters. A set of hyper parameters controls the desired amount of sparsity. Even though these approaches have shown promising performance, they often involve several user-defined hyperparameters and non-Gaussian distributions or discrete latent variables, which complicates the learning process.

In this paper, we adopt a different standpoint. We propose to structure the latent space via the usage of a sparsity-promoting dictionary model: each latent vector is assumed to have a sparse representation over a dictionary, i.e., it can be written as a weighted sum of only few columns of the dictionary. As a sparsity regularizer, we adopt a zero-mean Gaussian prior distribution with learnable variances: this is inspired by the relevance vector machines~\cite{tipping2001sparse}, which has proven effective for promoting sparsity. Furthermore, in contrast to non-Gaussian priors, it does not complicate the learning procedure, and does not introduce any extra sparsity-adjusting parameter. We derive an efficient variational inference scheme to learn the parameters, with a negligible extra computational burden compared to a plain VAE. Our experimental results on speech generative modeling demonstrate the effectiveness of the proposed model in promoting sparsity while preserving the reconstructed speech quality.

The rest of the paper is organized as follows. Section~\ref{sec:vae} reviews the standard VAE model and inference. Section~\ref{sec:prop} presents the proposed approach and its application to speech spectrogram modeling. Experiments are conducted in Section~\ref{sec:exp}. Finally, Section~\ref{sec:conc} concludes the paper.

\section{Background on VAE}
\label{sec:vae}

\subsection{Generative modeling}

Let $\vect{s}=\{\vect{s}_1,\ldots,\vect{s}_N\}$ denote a set of training data with ${\vect{s}_i \in \mathbb{R}^n}$. The core idea in a generative modeling context is to encode the data via some latent variables, denoted ${\vect{z}=\{\vect{z}_1,\ldots,\vect{z}_N\}}$ with ${\vect{z}_i \in \mathbb{R}^m}$, where usually ${m\ll n}$. Then, the goal is to model the joint distribution $p{(\vect{s}, \vect{z}) = p(\vect{s}| \vect{z})\cdot p(\vect{z})}$. To this end, the variational autoencoding framework \cite{KingW14} assumes some parametric forms for the generative distribution $p(\vect{s}| \vect{z})$, which is also referred to as the \emph{decoder}, and for the prior distribution $p(\vect{z})$, which are typically expressed as Gaussian distributions:
\begin{equation}
\begin{cases}
p_{\theta}(\vect{s}| \vect{z}) = \mathcal{N}(\bs{\mu}_{\theta}(\vect{z}), \diag(\bs{\sigma}_{\theta}^2(\vect{z})),\\
p(\vect{z}) = \mathcal{N}(\bs{0}, \vect{I}),
\end{cases}
\end{equation}
where $\mathcal{N}(\bs{\mu}, \bs{\Sigma})$ denotes a Gaussian distribution with mean $\bs{\mu}$ and covariance matrix $\bs{\Sigma}$, $\vect{I}$ denotes the identity matrix of appropriate size, and the power is applied element-wise. Furthermore,  $\bs{\mu}_{\theta}(.)$ and $\bs{\sigma}_{\theta}(.)$ are some non-linear functions denoting the mean and standard deviation, which are implemented via some deep neural networks (DNNs). To learn the model parameters, i.e., $\theta$, one would need to compute the posterior distribution of the latent codes, that is $p_{\theta}(\vect{z}| \vect{s})$, which is, unfortunately, intractable to compute. The VAE framework proposes to approximate this distribution with a parametric Gaussian distribution, called the \emph{encoder}, as follows:
\begin{equation}
	q_{\psi}(\vect{z}| \vect{s}) = \mathcal{N}(\bs{\mu}_{\psi}(\vect{s}), \diag(\bs{\sigma}_{\psi}^2(\vect{s})),
\end{equation}
where $\bs{\mu}_{\psi}$ and $\bs{\sigma}_{\psi}$ are also implemented using some DNNs with parameters $\psi$.

\subsection{Parameter estimation}
\label{sec:vae_inf}
The whole set of parameters $\Phi=\{\theta, \psi\}$ is learned using the following variational procedure. Instead of directly optimizing the data log-likelihood $\log p_{\theta}(\vect{s})$, which is intractable, a lower bound $\mathcal{L}$ is targeted such that $\mathcal{L}(\Phi;\vect{s}) \le \log p_{\theta}(\vect{s})$. To that end, the evidence lower bound (ELBO) of the data log-likelihood is considered, which is defined as:
\begin{equation}\label{eq:elbo}
\mathcal{L}(\Phi;\vect{s}) = \E_{q_{\psi}(\vect{z}| \vect{s}) }[\log p_{\theta}(\vect{s}| \vect{z})] - \mathcal{D}_{\textsc{kl}}(q_\psi(\vect{z}|\vect{s}) \|p(\vect{z})), 
\end{equation}
where $\mathcal{D}_{\textsc{kl}}(q \|p)$ represents the Kullback–Leibler (KL) divergence between $q$ and $p$. The KL divergence acts as a regularizer in the above formula, whereas the first term measures the reconstruction quality of the model. Then, the overall training phase consists of optimizing $\mathcal{L}(\Phi;\vect{s}) $ over $\Phi$ using a gradient-based optimizer together with the application of the \emph{reparametrization trick}, which enables backpropagation through the encoder parameters \cite{KingW14}.

\section{Proposed framework}
\label{sec:prop}

\subsection{Model}
\label{sec:model}

\begin{figure}[t]
 \centering
 \includegraphics[width=0.9\linewidth]{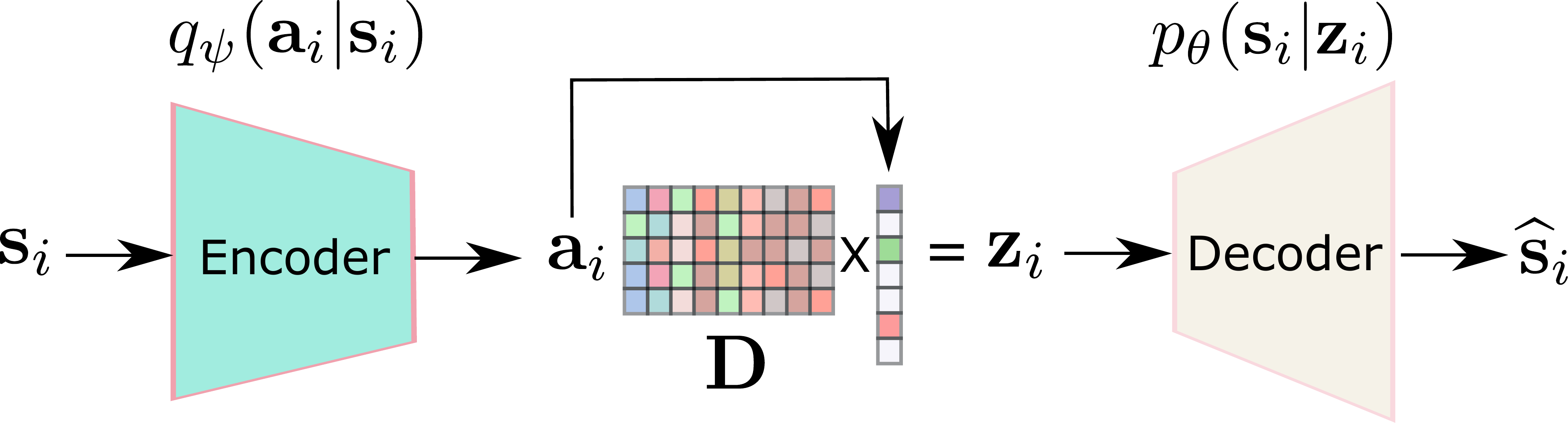}
 \vspace{-0.7em}
 \caption{Schematic diagram of the proposed VAE with a sparse dictionary model.}
 \label{fig:sdm_vae}
\end{figure}

The main idea underlying our proposal consists in structuring the latent space by considering a sparse dictionary model for each latent code $\vect{z}_i$ as follows:
\begin{equation}\label{eq:zmodel}
	\vect{z}_i = \vect{D}\vect{a}_i, \quad \forall i,
\end{equation}
where $ \vect{D}\in\mathbb{R}^{m\times k}$ is a dictionary with unit-norm columns (to avoid scale ambiguity), and $\vect{a}_i\in\Rbb^k$ denotes a (sparse) representation. In this work, we consider that $\vect{D}$ is a fixed dictionary, even though it could also be learned jointly with the code vector $\vect{a}_i$. Note that one way to achieve a sparser representation is to consider an overcomplete dictionary, that is, such that $k>m$. The proposed model is illustrated in Fig.~\ref{fig:sdm_vae}.

The motivation behind the model in~\eqref{eq:zmodel} is to keep the reconstruction quality of VAEs intact, while at the same time promoting interpretability of the latent space by enforcing sparsity of the code vector $\vect{a}_i$. To impose such a constraint, one possible approach is to leverage some sparsity-promoting prior, such as based on the Laplace distribution~\cite{Mohamed2012laplace}. However, such distributions involve non-smooth or complicated forms, which makes parameter inference and training more cumbersome. Instead, we propose to use the following zero-mean Gaussian prior:
\begin{equation}
p(\vect{a}_i;\bs{\gamma}_i) = \mathcal{N}(\bs{0}, \text{diag}(\bs{\gamma}_i)),
\end{equation}
in which $\bs{\gamma}_i$ is a vector of learnable variances. This prior has already proven successful in obtaining sparse solutions for linear regression and classification tasks \cite{tipping2001sparse,wipf2004sparse}. Indeed, as discussed in \cite{wipf2004sparse}, if one considers an inverse Gamma hyperprior on $\bs{\gamma}_i$ in a Bayesian setting, then the prior distribution, which is given~by:
\begin{equation}
p(\vect{a}_i) = \int p(\vect{a}_i|\bs{\gamma}_i)\cdot p(\bs{\gamma}_i) \text{d}\bs{\gamma}_i,
\label{eq:prior_a}
\end{equation}
becomes a Student's t-distribution, which has been shown to promote sparsity due to its sharp peak at zero~\cite{tipping2001sparse}. Even though we do not consider such a hyperprior explicitly, we still observe experimentally (see Section~\ref{sec:exp}) that the model given by~\eqref{eq:zmodel} and~\eqref{eq:prior_a} promotes sparsity. We refer the reader to~\cite{wipf2004sparse} for more detailed discussions.

\subsection{Inference}\label{sec:inference}
The inference phase follows the standard procedure of VAEs described in Section~\ref{sec:vae_inf}, with the main difference  being that the encoder is now used to encode each $\vect{s}_i$ into $\vect{a}_i$, then compute $\vect{z}_i$ via the dictionary model in \eqref{eq:zmodel}, and finally pass it to the decoder to reconstruct $\vect{s}_i$. Therefore, $\vect{z}_i$ only implicitly participates in the inference. Then, the cost function to optimize is:
\begin{align}
	\mathcal{L}(\Phi,\bs{\gamma};\vect{s}) &= \E_{q_{\psi}(\vect{a}| \vect{s}) }[\log p_{\theta}(\vect{s}| \vect{z})] - \mathcal{D}_{\textsc{kl}}(q_\psi(\vect{a}|\vect{s}) \|p(\vect{a}; \bs{\gamma})), \nonumber \\
	&\text{with} \quad  \vect{z}_i=\vect{D}\vect{a}_i \quad \forall i. \label{eq:elboa}
\end{align}
It should be noted that $\Phi$ are called \textit{amortized} parameters, as they are shared among all the training data, while ${\bs{\gamma}=\{\bs{\gamma}_1,\ldots,\bs{\gamma}_N\}}$ are parameters specific to each data point.

We use an alternating minimization approach to update $\Phi$ and $\bs{\gamma}$ by alternately fixing one variable and optimizing the cost function over the other one. This corresponds to solving the following two sub-problems iteratively:
\begin{subnumcases}{}
	\text{Update}~ \bs{\gamma}: & $\bs{\gamma} \leftarrow \argmin_{\bs{\gamma}}~\mathcal{D}_{\textsc{kl}}(q_\psi(\vect{a}|\vect{s}) \|p(\vect{a}; \bs{\gamma}))$ \label{eq:gamma}
	\\
	\text{Update}~ \Phi: & $\Phi \leftarrow \argmax_{\Phi}~\mathcal{L}(\Phi,\bs{\gamma};\vect{s})$. \label{eq:phi}
\end{subnumcases} 
Since the KL divergence between two Gaussian distributions can be obtained in closed-form, it is straightforward to solve~\eqref{eq:gamma}, which yields the following update rule:
\begin{equation}\label{eq:gammaup}
	\bs{\gamma}_i = \E_{q_{\psi}( \vect{a}_i| \vect{s}) }[\vect{a}_i^2]={\mu}_{\psi}^2(\vect{s}_i) + {\sigma}_{\psi}^2(\vect{s}_i), \quad \forall i.
\end{equation}
Then, we resort to the reparametrization trick~\cite{KingW14} to approximate the expectation in~\eqref{eq:elboa}, and thus reformulate sub-problem~\eqref{eq:phi}. Specifically, we approximate it using a single sample given by ${\vect{a}_i = \bs{\mu}_{\psi}(\vect{s}_i) + \bs{\sigma}_{\psi}(\vect{s}_i) \odot \bs{\epsilon}_i}$, where $\odot$ denotes the element-wise multiplication, and ${\bs{\epsilon}_i \sim \mathcal{N}(\boldsymbol{0}, \vect{I})}$. We then compute ${\vect{z}_i = \vect{D}\vect{a}_i}$ for each data sample $i$ in order to finalize computation of the following approximated loss:
\begin{equation}\label{eq:elboap}
\hat{\mathcal{L}}(\Phi,\bs{\gamma};\vect{s}) = \log p_{\theta}(\vect{s}| \vect{z}) - \mathcal{D}_{\textsc{kl}}(q_\psi(\vect{a}|\vect{s}) \|p(\vect{a}; \bs{\gamma})).
\end{equation}
Then, we optimize~\eqref{eq:elboap} by a single gradient ascent step, as in the standard VAE learning framework, which yields the update on~$\Phi$. The proposed procedure, called the sparsity-promoting dictionary model VAE (SDM-VAE) is summarized in Algorithm~\ref{alg:prop}. Note that for practical implementation, the training data is split into mini-batches, and a stochastic gradient algorithm is applied sequentially to each mini-batch.

\begin{algorithm}[t]
	\caption{SDM-VAE}\label{alg:prop}
	\begin{algorithmic}[1]
		\State \textbf{Input:} Training data $\vect{s}=\{\vect{s}_1,\ldots,\vect{s}_N\}$, $\vect{D}$.

		\State \textbf{Initialize:} $\Phi=\{\theta, \psi\}$ with random entries.
		\State \textbf{While} stopping criterion not met \textbf{do}:
		\State \textbf{For} each mini-batch $(b)$:
		\begin{itemize}
			[leftmargin=*]
			\compresslist
			\item \textbf{$\bs{\gamma}^{(b)}$ - update:} Using \eqref{eq:gammaup}
			\item \textbf{Reparametrization:}  
			\begin{itemize}
			    \item $\epsilon^{(b)} \sim \mathcal{N}(\boldsymbol{0}, \vect{I})$
			    \item ${\vect{a}^{(b)} = \bs{\mu}_{\psi}(\vect{s}^{(b)}) + \bs{\sigma}_{\psi}(\vect{s}^{(b)}) \odot \bs{\epsilon}^{(b)}}$
			\end{itemize}
			\item \textbf{$\vect{z}^{(b)}$ - update:} Using $\vect{z}^{(b)}=\vect{D}{\vect{a}^{(b)}}$
			\item \textbf{$\vect{\Phi}$ - update:} Using one-step gradient ascent on \eqref{eq:elboap} 
			
		\end{itemize}
		\State \textbf{End for}
		\State \textbf{End while}
		
	\end{algorithmic}
\end{algorithm}

\subsection{Application to speech modeling}\label{sec:app}

In this paper, we evaluate the effectiveness of the proposed approach for the modeling of speech signals, that is, speech \emph{analysis-resynthesis}~\cite{bie2021benchmark}. To that end, we first compute the short-time Fourier transform (STFT) of the time-domain signals, which yields the complex-valued data ${\mathbf{x} = \{ \mathbf{x}_1, \ldots, \mathbf{x}_N \}}$, where $N$ is the total number of time frames. In practice, the VAE processes and retrieves power spectrograms ${\vect{s} = |\vect{x}|^2}$. In terms of probabilistic modeling, this boils down to assuming that the data follows a circularly-symmetric zero-mean complex-valued Gaussian distribution, which is a common assumption when modeling speech signals with VAEs~\cite{leglaive2018variance}.

Once the VAE model is trained on the power spectrograms, we perform speech analysis-resynthesis on the test data, similarly to \cite{bie2021benchmark}. More precisely, once the input power spectrogram is encoded to obtain the posterior parameters, the latent's posterior mean is fed to the decoder to reconstruct the input spectrogram. Then, the magnitude spectrogram estimate $\hat{\mathbf{s}}^{1/2}$ is combined with the input STFT's phase $\angle \mathbf{x}$ in order to obtain a complex-valued STFT: ${\hat{\mathbf{x}} = \hat{\mathbf{s}}^{1/2} \odot \exp({j \angle \mathbf{x}})}$. Finally, time-domain estimates are retrieved by applying the inverse STFT to~$\hat{\mathbf{x}}$. Note that, as such, the VAE ignores the phase information, which is only exploited for re-synthesizing time-domain signals. We leave modeling the phase in VAEs~\cite{Nugraha2019vmvae} or extending the proposed method to complex-valued autoencoders, e.g., by overcoming the circularly-symmetric assumption~\cite{Magron2017anis}, to future work.

As for the dictionary, we consider a discrete cosine transform (DCT) matrix with $k$ atoms, that is, the $r$-th column of the DCT dictionary $\vect{D}$ is defined as follows~\cite{aharon2006k}:
\begin{equation}\label{eq:dct}
    \vect{d}_{r} = [\cos\big((\ell-1) \pi r/k\big)]_{\ell=1}^{m},
\end{equation}
followed by mean subtraction and normalization to ensure unit-norm columns.

\begin{table*}[t]
	\setlength{\tabcolsep}{0.6em} 
	\caption{Reconstruction quality and sparsity measure for various VAE-based methods in terms of PESQ, STOI, and Hoyer scores. The results are averaged over the test set (the standard deviation for all methods is in the order of $0.01$ for PESQ, and $0.001$ for STOI and the Hoyer score).}
	\centering 
	\aboverulesep = 0.25pt
    \belowrulesep = 0.25pt
	{\renewcommand{\arraystretch}{1.3}
	\begin{tabular}{l| l| c c c| c c c} 
		\toprule
		\multicolumn{2}{l|}{Dimension of $\vect{z}$} & \multicolumn{3}{c|}{$m=32$} & \multicolumn{3}{c}{$m=64$} \\
		\midrule
		\multicolumn{2}{l|}{} & PESQ & STOI & Hoyer & PESQ & STOI & Hoyer \\ 
		\midrule
		\midrule  
		 \multicolumn{2}{l|}{VAE \cite{leglaive2018variance}}  & 3.29 & 0.85 & 0.40 & 3.26 & 0.85 & 0.56\\
		 \midrule
		  \multirow{3}{*}[-0.55em]{VSC \cite{tonolini2020variational}}  & $\alpha = 0.05$ & 3.00 & 0.81 & 0.57 & 3.25 & 0.84 & 0.51\\
		  \cmidrule{2-8}
		     & $\alpha = 0.5$ & 3.25 & 0.84 & 0.54 & 3.32 & 0.85 & 0.65 \\
		     \cmidrule{2-8}
		     & $\alpha = 0.9$ & 3.25 & 0.84 & 0.47 & 3.26 & 0.85 & 0.60 \\
		     \midrule
		  \multirow{3}{*}[-0.3em]{SDM-VAE}  & $\vect{I}$ & {3.33} & {0.86} & {0.64} & \textbf{3.45} & \textbf{0.87} & 0.73\\
		  \cmidrule{2-8}
		  & DCT ($k=32$) & {3.37} & {0.86} & {0.66} & 3.28 & 0.84 & 0.66\\
		  \cmidrule{2-8}
		     & DCT ($k=64$) & 3.32 & 0.86 & \textbf{0.87} & 3.33 & 0.86 & 0.76\\
		\bottomrule 
	\end{tabular}}
	\label{table:speechrecons} 
\end{table*}

\section{Experiments}
\label{sec:exp}

In this section, we assess the potential of our proposed VAE model in terms of speech modeling. The code implementing the proposed VAE model is available online for reproducibility.\footnote{\url{https://gitlab.inria.fr/smostafa/sdm-vae}}

\subsection{Protocol}

\subsubsection{Data}
We use the speech data in the TCD-TIMIT corpus \cite{harte2015tcd} for training and evaluating the model. It includes speech utterances from 56 English speakers with an Irish accent, uttering 98 different sentences, each with an approximate length of 5 seconds, and sampled at 16 kHz. The total speech data duration is about 8 hours. 39 speakers are used for training, 8 for validation, and the remaining 9 speakers for testing. The STFT parameters are as follows. The STFT is computed with a 1024 samples-long (64 ms) sine window, 75$\%$ overlap and no zero-padding, which yields STFT frames of length $n=513$.

\subsubsection{Methods \& model architecture} \label{sec:exp_methods}
As baseline methods, we compare the performance of a standard VAE\footnote{\url{https://gitlab.inria.fr/smostafa/avse-vae}} \cite{leglaive2018variance}, the VSC model\footnote{\url{https://github.com/Alfo5123/Variational-Sparse-Coding}} \cite{tonolini2020variational}, and the proposed SDM-VAE model. All the VAE models follow the same {simple} encoder-decoder architecture as the one proposed in \cite{leglaive2018variance} {in order to focus our experiments on the latent space structure}. More precisely, both the encoder and decoder are fully-connected networks using a single hidden layer with 128 nodes and hyperbolic tangent activation functions. The dimension of the latent space is $m\in\{32, 64\}$. The dictionary in SDM-VAE is a DCT matrix, as defined in~\eqref{eq:dct}, which contains $k\in \{ 32, 64\}$ atoms. {These parameters values are chosen according to prior studies~\cite{leglaive2018variance,tonolini2020variational}, where they have shown good performance.} For comparison, we also consider the identity matrix as the dictionary $\vect{D} = \vect{I}$. In this setting, the model becomes similar to a classical VAE, except the latent prior now is $\vect{z}_i \sim \mathcal{N}(\bs{0}, \text{diag}(\bs{\gamma}_i))$ instead of the standard normal prior $\vect{z}_i \sim \mathcal{N}(\bs{0}, \vect{I})$.

\subsubsection{Parameter settings \& training}
All the VAE variants are implemented in PyTorch \cite{paszke2019pytorch}, and trained with mini-batch stochastic gradient descent using the Adam optimizer~\cite{KingmaB15} with a learning rate equal to $0.0001$, and a batch size of 128. As stopping criterion we use early stopping on the validation set with a patience of 20 epochs, meaning that the training stops if the validation loss does not improve after 20 consecutive epochs. For the sparsity parameter of the VSC model, denoted $\alpha$, we have experimented three different values: $\{ 0.05, 0.5,0.9\}$, where a lower value corresponds to a sparser representation.
\subsubsection{Evaluation} To evaluate the compared methods in terms of reconstruction quality, we compute the short-term objective intelligibility (STOI) measure~\cite{Taal2011stoi} and the perceptual evaluation of speech quality (PESQ) score~\cite{Rix2001pesq}. These metrics respectively belong to the $[-0.5,4.5]$ and $[0,1]$ range, and higher is better. We also compute the Hoyer metric~\cite{Hoyer2004sparse} in order to evaluate the sparsity of the latent codes. For a vector $\vect{z}\in \Rbb^m$, the Hoyer metric is defined as follows:
\begin{equation}
    \text{Hoyer}(\vect{z})=\frac{\sqrt{m}-\|\vect{z}\|_1/\|\vect{z}\|_2}{\sqrt{m}-1},
\end{equation}
where $\|.\|_1$ and $\|.\|_2$ respectively denote the $\ell_1$ and $\ell_2$ norms.
The Hoyer metric ranges in $[0,1]$, and a higher score corresponds to a more sparse latent feature. It should be emphasized that for the proposed SDM-VAE model, the sparsity is measured on $\vect{a}$, since it is the interpretable latent code in this case.

\subsection{Results}

From the results, averaged over all the test samples, and displayed in Table~\ref{table:speechrecons}, we can draw several conclusions. First, we observe that for the classical VAE, the PESQ decreases when going from $m=32$ to $64$. Overall, the opposite behavior is observed for the sparse VAE models, for which increasing the latent space dimension does not sacrifice the reconstruction quality (except for SDM-VAE when using a DCT dictionary with $k=32$ atoms, which is expected as the dictionary becomes undercomplete for $m=64$). These results confirm a known advantage of sparsity as a regularizer for the model to avoid overfitting.

For VSC, we see that increasing sparsity (i.e., decreasing~$\alpha$) results in sacrifying the reconstruction quality in terms of PESQ when $m=32$, although a different trend is observed for $m=64$. Our proposed SDM-VAE model using the DCT dictionary is less sensitive to the sparsity level, since it exhibits more stable STOI values, and an increasing PESQ when the sparsity score increases.

We also observe that increasing the number of atoms in the dictionary results in an increased sparsity score. In particular, the highest sparsity score is obtained for $m=32$ and when using an overcomplete dictionary ($k>m$). Nonetheless, this setting does not improve reconstruction quality compared to using a complete dictionary. As a result, using $k=m$ would be an appropriate and simple guideline, since it maximizes reconstruction quality and avoid fine-tuning the dictionary size.

Interestingly, we observe that the best SDM-VAE results are obtained when using an identity dictionary rather than the DCT. This shows that a simple change in the prior structure compared to the baseline VAE (as detailed in Section~\ref{sec:exp_methods}) is an efficient way to both improve performance and promote sparsity, rather than resorting to more involved models such as in VSC. Nevertheless, the DCT dictionary remains the best performing option when $m=32$. This motivates a future investigation on the design of more optimal dictionaries, or the joint learning of the dictionary along with other parameters based on dictionary learning algorithms from the literature~\cite{aharon2006k,sadeghi2020dictionary}.

Finally, we remark that SDM-VAE yields the best results in terms of both reconstruction quality (PESQ and STOI) and sparsity (Hoyer score). This outlines the potential of our proposal, which does not rely on a complicated inference scheme or extra-parameter tuning, and exhibits a stable performance while enabling interpretability and avoiding overfitting.

\section{Conclusions}\label{sec:conc}

In this work, we have proposed a novel approach for promoting sparsity in VAEs based on structuring the latent space using a sparse dictionary model. We derived a simple inference scheme, which does not require any fine-tuning of hyperparameters. Experiments conducted on speech signals modeling have demonstrated the potential of this technique compared to other existing sparse approaches. In particular, the proposed method improves speech reconstruction quality, and efficiently exploits sparsity to improve interpretability of the latent representation while reducing the risk of overfitting. Future work will focus on learning the dictionary along with other parameters, possibly with some mutual coherence constraint~\cite{sadeghi2020dictionary}. Extending the developed model to dynamical VAEs~\cite{bie2021benchmark}, and exploiting it in speech enhancement~\cite{leglaive2018variance} and source separation applications~\cite{nguyen2021deep} are other future directions to pursue.

\section{Acknowledgements}
Experiments presented in this paper were carried out using the Grid'5000 testbed, supported by a scientific interest group hosted by Inria and including CNRS, RENATER and several Universities as well as other organizations (see https://www.grid5000.fr). 

\balance
\bibliographystyle{IEEEtran}
\bibliography{mybib}

\end{document}